\documentclass[11pt]{article}

\usepackage{acl}

\usepackage{times}
\usepackage{latexsym}

\usepackage[T1]{fontenc}

\usepackage[utf8]{inputenc}
\usepackage{amsmath} 
\usepackage{amssymb} 

\usepackage{algorithm}
\usepackage{algpseudocode}

\usepackage{multirow}
\usepackage{tabularx}
\usepackage{xcolor}
\usepackage{booktabs}
\usepackage{microtype}

\usepackage{inconsolata}

\usepackage{graphicx}
\usepackage{booktabs}

\title{Is the Future Predictable? Modeling Future Predictability for Reliable Temporal Reasoning}

\author{
 \textbf{Yongyi Huang\textsuperscript{1}},
 \textbf{Huaping Zhang\textsuperscript{1}},
 \textbf{Wenyao Cui\textsuperscript{1,2}},
 \textbf{Qiuchi Li\textsuperscript{1}},
\\
 \textbf{Lei Li\textsuperscript{1}},
 \textbf{Baohua Zhang\textsuperscript{1}\thanks{Corresponding author}}
\\
 \textsuperscript{1}Beijing Institute of Technology,
 \textsuperscript{2}Zhongguancun Academy
\\
 \small{
   \textbf{Correspondence:} \href{mailto:huangyy\_ov0@163.com}{huangyy\_ov0@163.com};
   \href{mailto:bhzhang\_bit@163.com}{bhzhang\_bit@163.com}
 }
}

\begin{document}
\maketitle
\begin{abstract}

Large language models (LLMs) remain unreliable when reasoning about the future.
Existing models often overlook the difference among future states, committing to specific answers even when the queried facts are inherently volatile.
To capture this challenge, we introduce \textbf{ChronoRegime}, a future-oriented temporal reasoning task that requires models to determine whether historical temporal evidence justifies commitment to a future answer. Furthermore, we propose \textbf{C}onfidence-Aware \textbf{T}emporal \textbf{R}easoning (\textbf{CTR}), a framework that estimates future predictability from temporal dynamics and regulates answer commitment accordingly.
Rather than performing unconditional forecasting, CTR commits only when the reconstructed evidence supports the target future state.
Across multiple LLMs, CTR improves strict overall accuracy by 6.7 percentage points and reduces hallucination rates by 18.4 percentage points on average over the plain QA baseline.
These results demonstrate that explicit predictability assessment is an effective mechanism for safer and more reliable temporal reasoning under future uncertainty.

\end{abstract}

\section{Introduction}

\begin{figure}[htb]
\centering
\includegraphics[width=1\columnwidth]{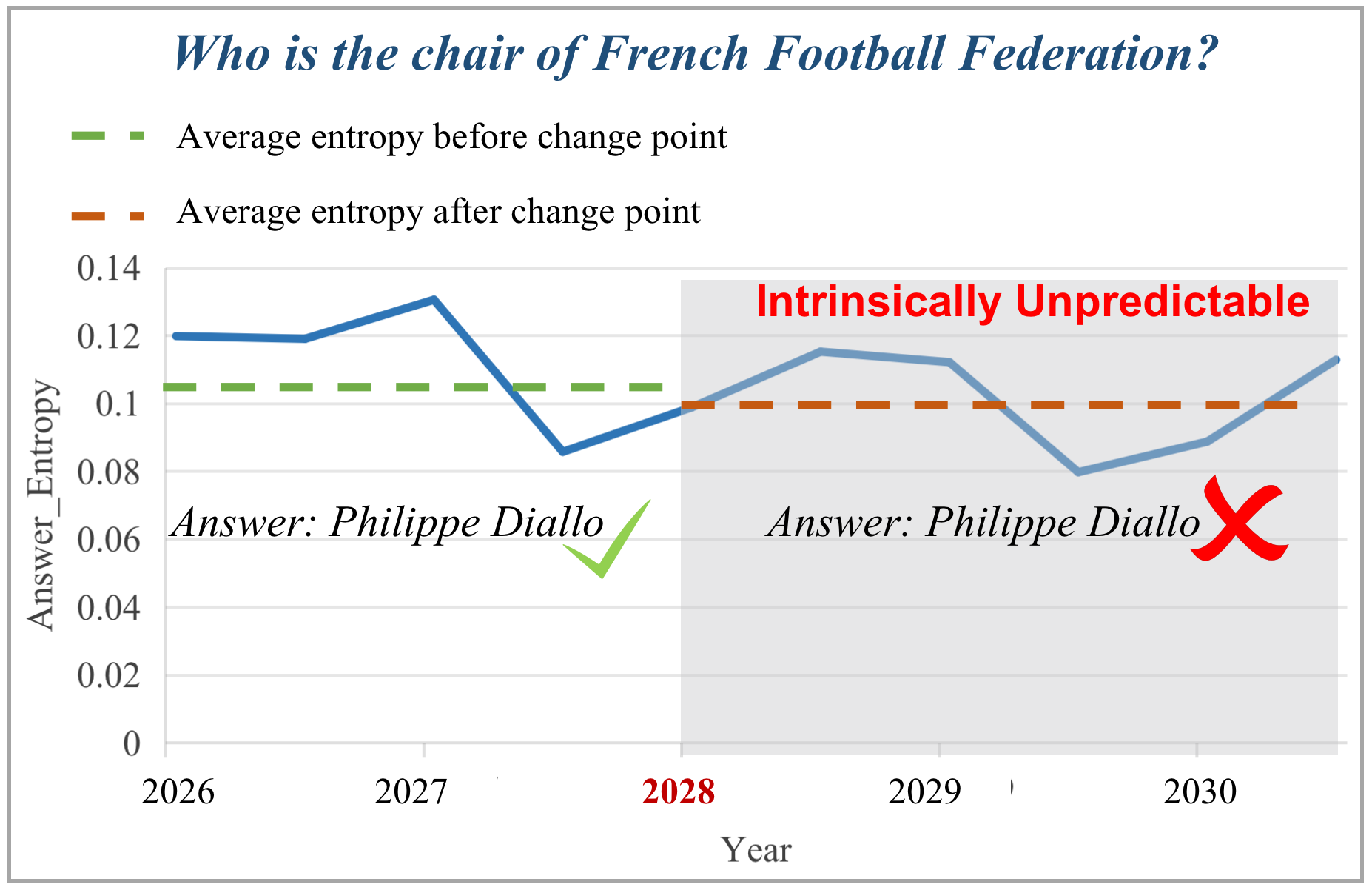} 
\caption{
An example of how LLaMA3.1 performs in future-oriented temporal reasoning tasks. The model is unable to distinguish the predictable boundaries, and gives confident but incorrect answers at times that are inherently unpredictable.
}
\label{fig:problem}
\end{figure}

Large language models (LLMs) have demonstrated impressive capabilities in factual reasoning and question answering~\citep{fatemi2024test,wang2024factuality,hu2025can}.
However, their reliability on time-sensitive queries remains a fundamental and largely unresolved challenge within temporal reasoning~\citep{dhingra2022time,wei-etal-2023-menatqa}.
Although advanced knowledge-augmentation strategies mitigate factual inconsistencies by grounding outputs in external knowledge~\citep{zhao2024set,bajpai2024temporally}, 
these systems remain highly susceptible to overconfident yet erroneous predictions when historical evidence is insufficient or the queried event is inherently unpredictable~\citep{wen2025know, zhou2026silence}. Such failures can have concrete real-world consequences: unsupported predictions about election outcomes may mislead users, while overconfident forecasts of disease progression or treatment response may distort clinical decision-making.

\begin{figure}[h]
\centering
\includegraphics[width=1\columnwidth]{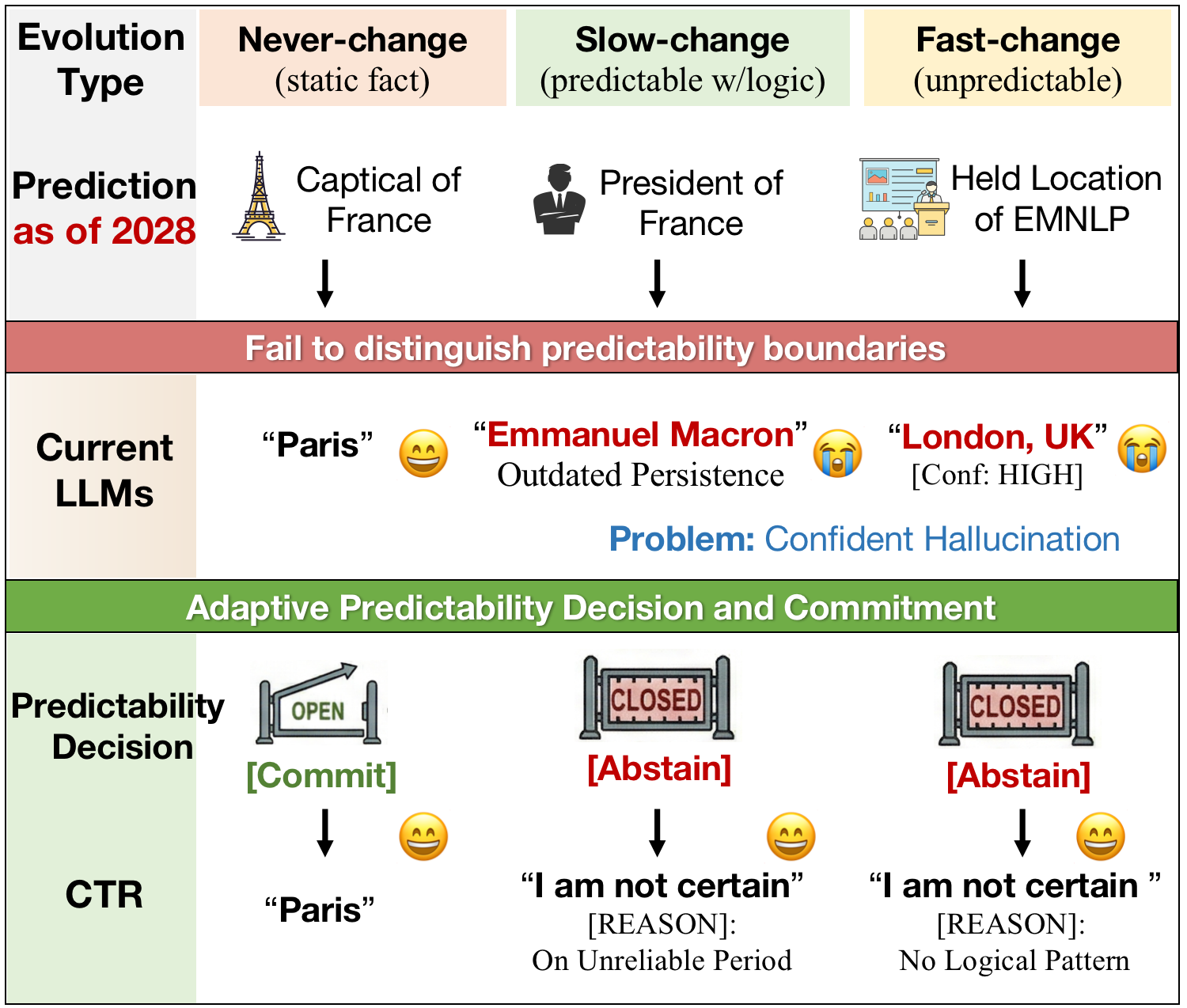} 
\caption{
Challenge of Current LLMs on future predictability modeling and the motivation of ChronoRegime in highlighting the needs for decision-making beyond unconditional prediction.
}
\label{fig:motivation}
\end{figure}

This reliability gap becomes especially acute in future-oriented temporal reasoning, where the predictability of the target fact varies across queries, time horizons, and temporal evolution patterns~\citep{vu2024freshllms,wu2024time,zeng2025futurex}.
For instance, as shown in Figure~\ref{fig:problem}, historical evidence may justify predicting \textit{"Philippe Diallo"} as the chair of the French Football Federation before 2028, but after the change point, the same historical evidence does not determine who will occupy the position in the future. 
More broadly, future queries span different evolution regimes, as illustrated in Figure~\ref{fig:motivation}: an unchanging fact such as the \textit{"capital of France"} can be predicted reliably; whereas a fast-changing query such as \textit{"Where will EMNLP be held as of 2028?"} may remain intrinsically unpredictable even with complete historical signals.


However, our preliminary study in Section~\ref{sec:prestudy} shows that current LLMs still struggle in future-oriented temporal reasoning, often committing to a specific future answer even when the underlying temporal evidence does not justify such commitment (e.g., predicting \textit{"London, UK"} for the location of EMNLP in 2028).
This persistent drop, together with uniformly weak boundary-aligned commitment behavior across models, highlights a critical gap in time-sensitive forecasting: the inability to assess whether available evidence justifies a factual commitment.

Motivated by these findings, we identify a distinct and more challenging task in future-oriented temporal reasoning: \textbf{ChronoRegime}. 
We define ChronoRegime as the ability to infer the predictability regime of a future fact from historical temporal signals and to calibrate prediction commitment accordingly.
Rather than merely producing a plausible future answer, the task asks whether the queried future state is predictable enough to warrant prediction.


We further propose \textbf{Confidence-Aware Temporal Reasoning (CTR)}, a framework for predictability-aware future factual reasoning. Rather than directly committing to an answer, CTR first estimates whether the queried fact lies in a predictable regime by modeling temporal evolution patterns in historical evidence, and only then decides whether to answer or abstain. This design shifts the task from unconditional forecasting to evidence-bounded commitment.


We evaluate CTR on updated versions of FRESHQA~\citep{vu2024freshllms} and TEMPLAMA~\citep{dhingra2022time}, using multiple LLM backbones, including LLaMA-3.1, Qwen3, and GPT-4o. CTR consistently improves future factual reasoning across models and substantially reduces overconfident failures on volatile queries, lowering hallucination rates by 18.4 percentage points on average. These results suggest that explicit predictability assessment provides an effective mechanism for calibrating model commitment under temporal uncertainty.

In summary, our contributions are as follows:
\begin{itemize}

    \item We identify future predictability modeling as a central challenge in future-oriented temporal reasoning and introduce \textbf{ChronoRegime}, a task that evaluates whether historical temporal evidence justifies commitment to a future prediction.
    
    \item We empirically show that current LLMs fail on future-oriented questions not merely due to missing or outdated knowledge, but because they cannot distinguish predictable from volatile future facts.
    
    \item We propose \textbf{CTR}, a predictability-aware framework that models temporal evolution signals and regulates answer commitment for more reliable temporal reasoning under future uncertainty.

\end{itemize}

\section{Related Work}

Temporal reasoning has long been a central topic in the study of large language models~\citep{Vashishtha2020TemporalRI,hu2023large,gurnee2023language,beniwal2024remember}. A substantial body of work has evaluated LLMs' temporal reasoning capabilities from multiple perspectives, leading to the development of benchmark datasets and evaluation metrics that assess factual correctness~\citep{fatemi2024test,chu2024timebench,zhu2025evolvebench,kasai2023realtime}, consistency~\citep{bajpai2024temporally, jia2024tiq}, and reasoning traceability~\citep{Xu2023SearchintheChainTT,bazaga2025learning,zhu2025evolvebench}. These studies consistently show that temporal factuality in LLMs is fundamentally challenged by the static and temporally mixed nature of pretraining corpora, which often results in outdated predictions, averaging of conflicting facts across time, and poorly calibrated confidence when models are queried about future states~\citep{dhingra2022time,vu2024freshllms,bajpai2024temporally,kasai2023realtime}.

One line of research improves LLMs’ temporal awareness through pretraining or fine-tuning strategies that incorporate explicit temporal signals, with the goal of enhancing temporal scope understanding and confidence calibration~\citep{ning2018joint, dhingra2022time, yang2023once, xiong2024large}. These approaches typically treat time as a conditioning variable that constrains the validity of factual knowledge. 
Another line of work enhances temporal reasoning through structured representations, most notably knowledge graph–based approaches that explicitly encode entities, relations, and temporal constraints~\citep{vu2024freshllms, xiong2024large, wang2024tram,hu2025can}. 
But they typically rely on predefined schemas, which makes graph updates costly and complex in dynamic settings where facts continuously evolve, thereby limiting their practicality for future-oriented reasoning that requires timely knowledge updates.

Another prominent direction augments LLMs with external retrieval mechanisms to refresh factual knowledge and reduce staleness~\citep{vu2024freshllms,wu2024time}. Retrieval-augmented and deep search methods leverage web search, document retrieval, or similar tools to improve performance on time-sensitive queries\citep{Xu2023SearchintheChainTT,Fu2024FastLearnAR}. 
However, most RAG-style pipelines implicitly treat retrieved documents as reliable evidence, despite differences in publication time and potential conflicts across sources.
The lack of explicit modeling of temporal validity and uncertainty allows errors to propagate through reasoning chains, resulting in hallucinations and inconsistencies~\citep{trivedi2023interleaving,vu2024freshllms,singal2024evidence,bajpai2024temporally}. 



Existing studies have substantially advanced temporal reasoning in LLMs, especially in temporal factuality evaluation, temporal adaptation, structured knowledge modeling, and retrieval-based knowledge refreshing.
By comparison, future-oriented temporal reasoning remains underexplored: how a model should reason about future states from historical temporal evidence under uncertainty is still not well understood.

\section{Preliminary Study}
\label{sec:prestudy}

We first evaluate current LLMs' ability to predict future facts from historical evidence through a controlled boundary-aligned probing study.

\begin{figure}[htb]
\centering
\includegraphics[width=1\columnwidth]{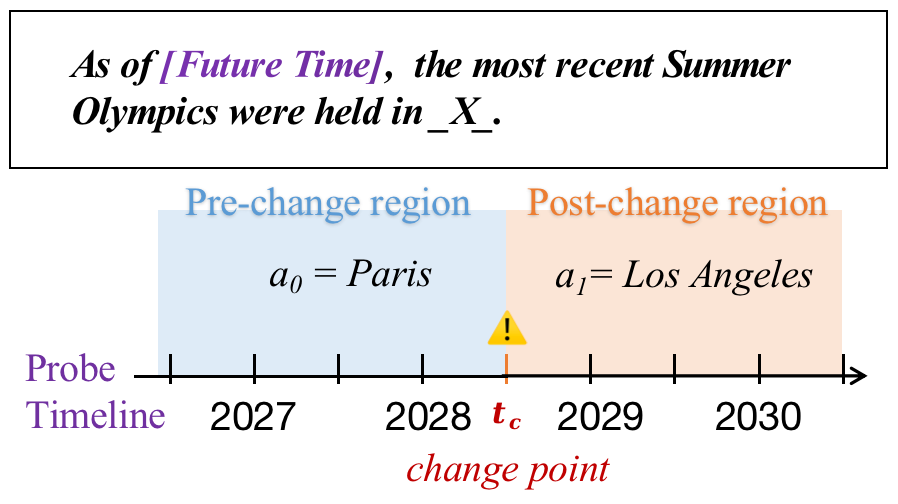} 
\caption{
Illustration of the boundary-probing setup. The change point acts as a predictability boundary: before it, the anchor answer is supported by temporal evidence; after it, the model must revise its commitment or avoid unsupported prediction.
}
\label{fig:prelim_setup}
\end{figure}

The setup involves a factual query on a future time, whose answer varies with respect to a certain time point, referred to as $change point$. Prior to the change point, the answer is definite and well supported with sufficient evidence from existing knowledge; after the change point, the same source of knowledge becomes insufficient for determining the same answer. For example, as shown in Figure~\ref{fig:prelim_setup}, one is asked to predict the location of most recent Summer Olympics. As of year 2028, one can deduce the correct answer with high confidence based on the information of the current Summer Olympics and the period of the event. After 2028, however, the answer to the same question becomes inherently indeterminate without additional information on the "next Summer Olympics". 

This study evaluates whether LLM outputs exhibit appropriate commitment behavior when target times move from an evidence-supported region to a region where the same evidence no longer justifies a concrete prediction. 
A reliable model should provide confident answers when evidence is sufficient, but revise, abstain, or express greater uncertainty once the evidence no longer supports a prediction.

\subsection{Problem Setting}

For evaluation, each instance consists of a factual query, an anchor answer \(a_0\) at the present time as a representation of the existing knowledge, a change point\(t_c\) as the transition boundary, and a post-boundary answer \(a_1\) when available.

We probe the model at a series of time stamps along the "future time line", spanning both pre-change and post-change regions.
At each probe, the model is prompted with the time-conditioned query for the target timestamp.
The annotated change point \(t_c\) and the post-boundary answer \(a_1\) are hidden from the model and used only for evaluation.


In our experiments, probes are placed at half-year offsets within a three-year window around the change point, and the boundary-aligned probe is counted as post-change.
The probing instances are drawn from the curated temporal QA pool described in Section~\ref{sec:datasetting}.

\subsection{Evaluation Metrics}

We evaluate boundary-aligned temporal reasoning with a compact set of metrics, and provide their formal definitions in Appendix~\ref{app:eval}.

\textbf{Pre- and Post-change Accuracy (Pre/Post Acc)} measure answer correctness before and after the change point, respectively.

\textbf{Change-aware Success (CAS)} evaluates whether the model correctly switches its commitment across the nearest change probes.

\textbf{Post-change Persistence Rate (Post-PR)} measures outdated persistence, i.e., whether the model continues to repeat the expired anchor answer after the boundary.

\textbf{Wrong Concrete Commitment Rate (WCCR)} measures wrong concrete commitment by counting concrete non-abstention predictions that are wrong.



\textbf{Entropy Shift ($\Delta H$)} denotes the post-minus-pre shift in answer entropy, capturing whether model uncertainty increases after the anchor answer loses validity.

\subsection{Analysis}

\begin{table}[htb]
\centering
\small
\setlength{\tabcolsep}{0.8pt}
\begin{tabular}{lcccccc}
\toprule
Model & PreAcc $\uparrow$ & PostAcc $\uparrow$ & CAS $\uparrow$ & Post-PR $\downarrow$ & WCCR $\downarrow$ & $\Delta H$  \\
\midrule
LLaMA & 0.500 & 0.158 & 0.000 & 0.429 & 0.832 & 0.006 \\
Qwen3    & \textbf{0.911} & 0.219 & 0.250 & 0.924 & 0.786 & 0.002 \\
Mistral  & 0.333 & 0.082 & 0.036 & \textbf{0.276} & 0.867 & 0.001 \\
GPT-4o   & 0.839 & \textbf{0.316} & \textbf{0.286} & 0.689 & \textbf{0.643} & \textbf{0.022} \\
\bottomrule
\end{tabular}
\caption{
Boundary-probing results.
}
\label{tab:preliminary}
\end{table}

\begin{figure}[htb]
\centering
\includegraphics[width=1\columnwidth]{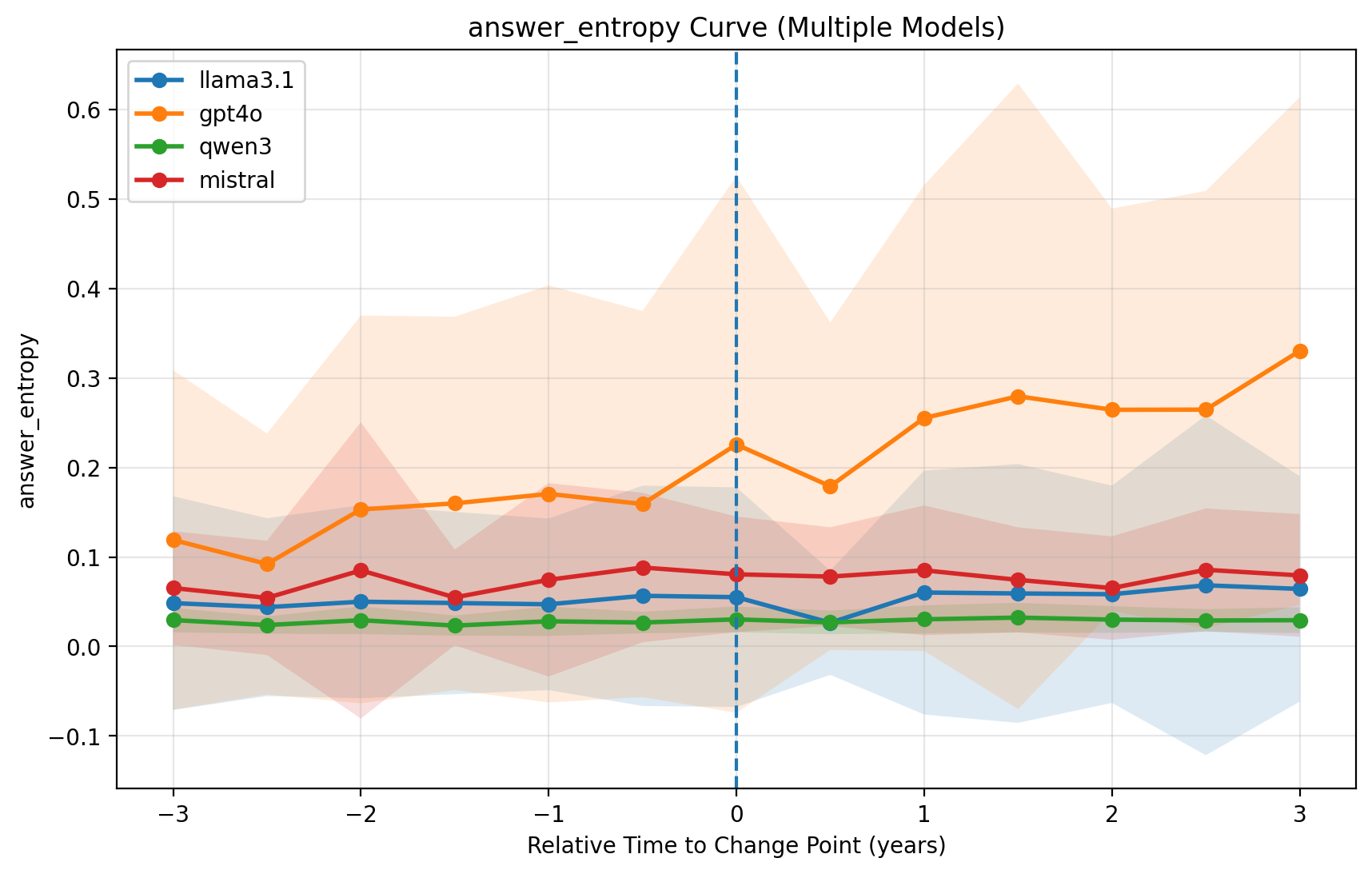} 
\caption{
Average answer entropy around factual change points.
Across models, uncertainty remains largely flat before and after temporal transitions, indicating that current LLMs do not reliably increase uncertainty when future commitment becomes less justified.
}
\label{fig:entropy-shift}
\end{figure}

Table~\ref{tab:preliminary} shows that current LLMs can often produce the anchor fact for pre-boundary target times, but fail to adjust their commitment once the queried target time falls beyond the annotated support boundary. Qwen3 and GPT-4o achieve high PreAcc scores of 0.911 and 0.839, respectively, yet their PostAcc drops to 0.219 and 0.316. CAS is also low across models, with the best result only 0.286 from GPT-4o, indicating weak boundary-aware switching.

Post-change failures are mostly unsafe commitments rather than cautious abstentions. WCCR remains high across models, ranging from 0.643 to 0.867. Among their wrong concrete commitments, Qwen3 and GPT-4o further show strong outdated persistence, repeating the expired anchor answer in 92.4\% and 68.9\% of post-change probes. Mistral has lower persistence but the highest WCCR, suggesting unsupported alternative guesses. \(\Delta H\) remains close to zero, showing that uncertainty barely rises after the boundary. 

Figure~\ref{fig:entropy-shift} further shows that model uncertainty remains largely flat around the annotated transition boundary.
The entropy shift $\Delta H$ is close to zero across all models, suggesting that they do not become meaningfully less confident when the anchor answer loses validity. This result indicates that current LLMs fail to infer the loss of evidential support from the time-conditioned query alone.

These findings motivate the central design of ChronoRegime and CTR. Reliable future-oriented reasoning requires more than predicting a plausible future answer; it requires estimating whether the available temporal evidence is sufficient to justify commitment.

\section{Methodology}
\begin{figure*}[htbp]
\centering
\includegraphics[width=\textwidth]{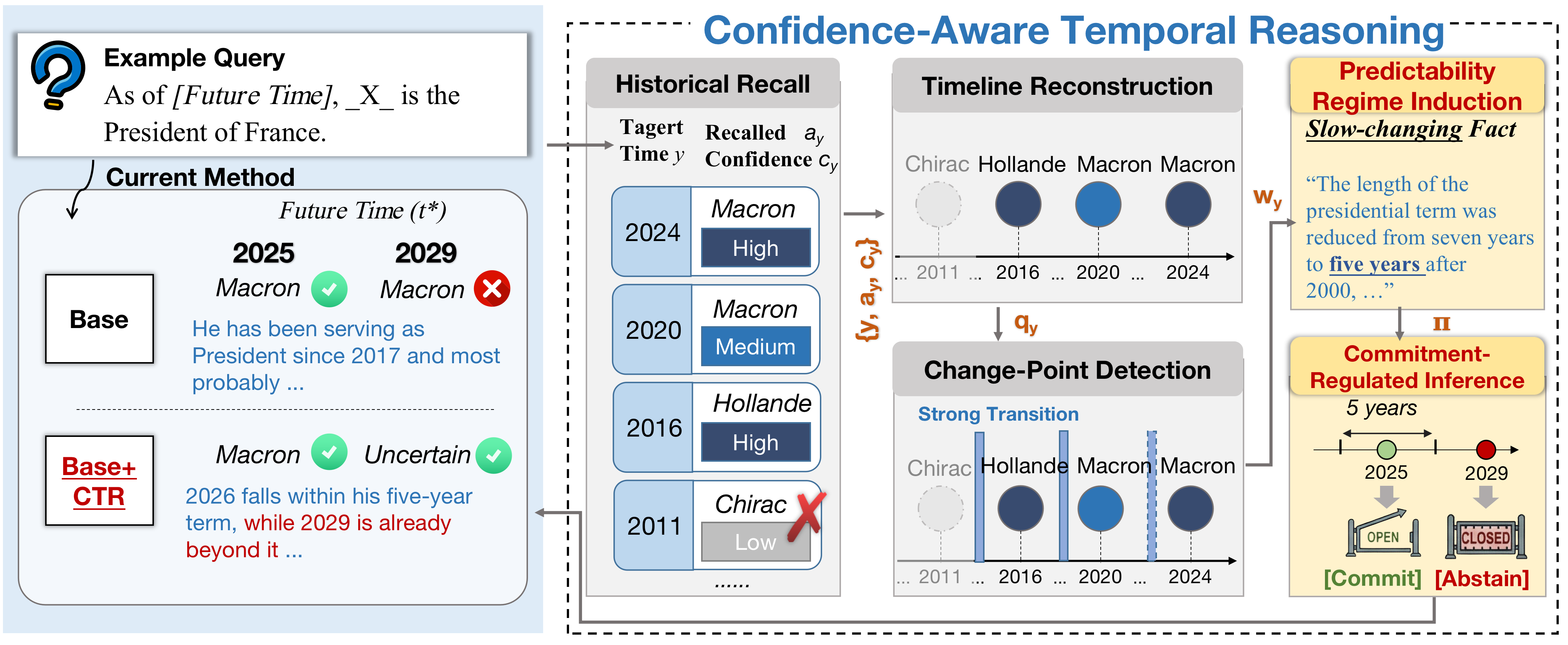}
\caption{\textbf{Overview for Confidence-Aware Temporal Reasoning (CTR).} 
CTR leverages entropy-based confidence to shift from static retrieval to dynamic evolution modeling. 
Notably, its confidence-aware mechanism suppresses erroneous historical evidence (e.g., the low-confidence 2011 entry) to mitigate error propagation. 
By inducing explicit evolution patterns, CTR overcomes the persistence bias seen in the traditional reasoning modes and prioritizes reliability, avoiding hallucinated extrapolation for volatile future queries.}
\label{fig:ctr}
\end{figure*}

We introduce \textbf{Confidence-Aware Temporal Reasoning (CTR)}, a framework for predictability-aware future factual reasoning under the \textbf{ChronoRegime} formulation. CTR follows a simple principle: before committing to a future answer, the model should first estimate whether the answer is predictable from temporal evidence. The overall architecture is illustrated in Figure~\ref{fig:ctr}.

\subsection{ChronoRegime Formulation}

Given a factual query $q$ and a future time $t^\ast$, ChronoRegime asks whether the future factual state ${a}_{t^\ast}$ is predictable from bounded historical temporal evidence. We define predictability in an evidence-bounded sense: a future state is predictable if the available temporal evidence supports a concrete answer with sufficient reliability; otherwise, the model should avoid commitment.
To operationalize this decision, we distinguish between two levels of temporal reasoning. First, the model infers a factual evolution regime $z \in \{\textsc{FAST}, \textsc{SLOW}, \textsc{NEVER}\}$, where $z$ describes the volatility profile of the queried fact. Second, the model maps this evolution regime, together with the reconstructed historical timeline and target time $t^\ast$, into a predictability decision $u \in \{\textsc{Predictable}, \textsc{Unpredictable}\}$.
The final output is therefore $\langle r, z, u, \hat{a}_{t^\ast} \rangle$, where $r$ is a reasoning trace explaining the decision and $\hat{a}_{t^\ast}$ is the predicted fill-in value for $q$ as of year $t^\ast$.

ChronoRegime treats predictability as an evidence-bounded decision \(u\): the model should commit only when the inferred regime, historical evidence, and target time jointly support a concrete answer. The benchmark-specific mapping from evolution regimes to predictability labels is described in Section~\ref{sec:datasetting}


\subsection{CTR Architecture}

CTR deliberately adopts conservative heuristics rather than optimized predictors. Its goal is not to maximize change-detection recall, but to regulate future commitment under uncertainty and avoid unfounded speculation when temporal evidence is weak or inconsistent.

Driven by this principle, CTR is formulated as a confidence-driven decision pipeline that progressively transforms noisy historical signals into a structured predictability regime for downstream commitment control. 
The framework consists of five-stage process: (1) Historical Recall; (2) Timeline Reconstruction; (3) Change Point Detection; (4) Predictability Regime Induction and (5) Commitment-Regulated Inference. 

We empirically examine whether these stages form a coherent decision
process, rather than isolated heuristic add-ons, through component
ablations in Section~\ref{sec:ablation}.


\paragraph{Historical Recall}

Given a future time $t^\ast$, CTR performs fine-grained historical probing over a window of past years:
\begin{equation}
    \mathcal{Y} = \{t^\ast - W, \ldots, t^\ast - 1\}.
\end{equation}
For each target year $y \in \mathcal{Y}$, the model is queried to produce a predicted fill-in value $a_y$ along with its token-level uncertainty, measured by the average generation entropy $H_y$. 

To obtain a calibrated reliability signal, we transform the raw entropy into a base confidence score $c_y$ via a piecewise linear mapping $f(H_y)$. This transformation normalizes the entropy within a predefined uncertainty range $[H_{\min}, H_{\max}]$ and inverts it:

\begin{equation}
c_y = 
\begin{cases} 
1, & H_y < H_{\min} \\
1 - \frac{H_y - H_{\min}}{H_{\max} - H_{\min}}, & H_{\min} \leq H_y \leq H_{\max} \\
0, & H_y > H_{\max}
\end{cases}
\end{equation}

where we empirically set $H_{\min} = 0$ and $H_{\max} = 0.05$. The resulting sequence of historical recall probes is represented as:
\begin{equation}
\mathcal{R} = \{(y, a_y, c_y)\}_{y \in \mathcal{Y}}
\end{equation}

This sequence $\mathcal{R}$ serves as the foundational evidence for subsequent timeline reconstruction and change point detection, where $c_y$ acts as the intrinsic weight for each temporal observation.


\paragraph{Timeline Reconstruction}

The recall probes are organized into a temporal timeline
\begin{equation}
    \mathcal{T} = \{(y, v_y, q_y)\},
\end{equation}
where $v_y = a_y$ denotes the recalled factual state at time $y$, and $q_y$
is a corrected reliability weight.

Importantly, $q_y$ is not directly inherited from the entropy-based confidence $c_y$.
Instead, CTR applies a probe-quality correction function that down-weights unreliable probes, such as those exhibiting abnormally long outputs, repetitive generation patterns, or malformed responses.
Formally, we define
\begin{equation}
    q_y = \textsc{Clip}_{[0,1]}\big(c_y \cdot (1 - \rho_{\text{len}}(y))
    \cdot (1 - \rho_{\text{rep}}(y))\big),
\end{equation}
where $\rho_{\text{len}}$ and $\rho_{\text{rep}}$ denote mild penalties based on
output length and $n$-gram repetition, respectively.
This correction reflects the fact that historical recall itself may be noisy
and should not be treated as uniformly reliable.

\paragraph{Change Point Detection}
Given the reconstructed timeline $\mathcal{T}$, CTR aims to identify years in which the factual state undergoes a genuine transition. We adopt a procedure that combines candidate proposal with consistency-aware reweighting.

\textit{Candidate proposal:}
We first obtain a set of candidate change-point years $\tilde{\mathcal{C}}$
by prompting the model to inspect the timeline and identify major state transitions.
If the model output is unavailable or invalid, we fall back to a deterministic rule that proposes time $y$ whenever $v_y \neq v_{y-1}$.
All candidates are restricted to years explicitly appearing in the timeline.

\textit{Consistency-aware reweighting:}
Each candidate $y \in \tilde{\mathcal{C}}$ is then filtered and assigned a confidence weight. Candidates with no effective state change ($v_y = v_{y-1}$) are discarded. For the remaining candidates, we compute
\begin{equation}
    w_y = \sqrt{q_{y-1}\, q_y} \cdot g_y,
\end{equation}
where $w_y$ denotes the confidence of the change point at time $y$ and
$g_y \in (0,1]$ penalizes isolated single-year deviations.

Specifically, a candidate is treated as singleton noise if it is unsupported by its immediate temporal context, i.e., $v_{y-1}=v_{y+1}$ but $v_y\neq v_{y-1}$, and its probe reliability is substantially lower than that of neighboring years.

In such cases, $g_y$ is reduced whenever
\begin{equation}
    q_y < \tau_{\text{single}} \cdot \min(q_{y-1}, q_{y+1}),
\end{equation}
thereby enforcing a local temporal consistency constraint.
This design favors transitions that persist across time and suppresses spurious one-off fluctuations.

The resulting set of weighted change points is denoted as
\begin{equation}
    \mathcal{C} = \{(y, w_y)\}.
\end{equation}

\paragraph{Predictability Regime Induction}

Given the reconstructed timeline $\mathcal{T}$ and weighted change points $\mathcal{C}$, CTR induces a high-level temporal evolution pattern by aggregating volatility statistics over the timeline.
Based on the frequency and persistence of state changes, the evolution regime $z$ is inferred.
Then, CTR maps $z$, together with the target time $t^\ast$ and the reconstructed support in $\mathcal{T}$, into a predictability decision $u$. 
The induced pattern representation is therefore

\begin{equation}
    \pi = \langle z, u,\text{key\_evidence} \rangle,
\end{equation}
which serves as an explicit evolution hypothesis for subsequent inference.

\paragraph{Commitment-Regulated Inference}
CTR performs future inference conditioned on the induced pattern representation $\pi$, where the inferred evolution type serves as the ChronoRegime decision signal for regulating downstream commitment. In this way, CTR first determines whether the future fact is predictable under the inferred temporal regime, and only then decides what answer to output.

When the predictability decision supports commitment, the model predicts the factual state at the future time $t^\ast$ as
\begin{equation}
    \hat{a}_{t^\ast} = \arg\max_a p(a \mid q, t^\ast, \pi),
\end{equation}
subject to strict constraints that prohibit unsupported extrapolation.


\section{Experiments}

\begin{table*}[t]
\centering
\small
\setlength{\tabcolsep}{6pt}
\begin{tabular*}{\textwidth}{@{\extracolsep{\fill}}llcccccc}
\toprule
\multirow{2}{*}{\textbf{Model}} &
\multirow{2}{*}{\textbf{Variant}} &
\multicolumn{4}{c}{\textbf{Answer Accuracy} (\%) $\uparrow$} &
\textbf{Type Acc.} (\%) $\uparrow$ &
\textbf{Halluc.} (\%) $\downarrow$ \\
\cmidrule(lr){3-6}
 & & Fast & Slow & Never & Overall &  &  \\
\midrule

LLaMA3.1
 & Base
 & \textbf{6.9} & 21.0 & 50.0 & 26.7 & 23.7 & 36.8 \\
 & +Controlled Retrieval
 & 5.9 & 29.4 & 53.4 & 30.6 & 26.1 & 33.2 \\
 & +CTR
 & \textbf{6.9} & \textbf{31.9} & \textbf{56.0} & \textbf{32.6}
 & \textbf{55.3} & \textbf{22.6} \\
\midrule

Qwen3
 & Base
 & 4.9 & 26.9 & 41.4 & 25.2 & 35.0 & 56.7 \\
 & +Controlled Retrieval
 & \textbf{8.8} & 21.0 & 47.4 & 26.4 & 54.9 & 54.6 \\
 & +CTR
 & 6.9 & \textbf{32.8} & \textbf{57.8} & \textbf{33.5}
 & \textbf{57.5} & \textbf{25.5} \\
\midrule

Mistral
 & Base
 & 7.8 & 12.6 & 50.0 & 24.0 & 29.7 & 40.1 \\
 & +Controlled Retrieval
 & 4.9 & 16.0 & 42.2 & 21.7 & 51.9 & 36.8 \\
 & +CTR
 & \textbf{8.8} & \textbf{21.0} & \textbf{50.9} & \textbf{27.6}
 & \textbf{52.8} & \textbf{18.4} \\
\midrule

GPT-4o
 & Base
 & 16.7 & 40.3 & 74.1 & 44.8 & 48.1 & 17.8 \\
 & +Controlled Retrieval
 & 13.7 & 46.2 & 79.3 & 47.8 & \textbf{63.8} & 13.7 \\
 & +CTR
 & \textbf{18.6} & \textbf{52.9} & \textbf{85.3} & \textbf{53.7}
 & \textbf{63.8} & \textbf{11.3} \\
\bottomrule
\end{tabular*}
\caption{
Main results (\%) on answer accuracy, type accuracy, and hallucination rate. Controlled Retrieval reasoning yields inconsistent gains over plain QA,
while CTR consistently improves performance on slow- and never-changing facts
and reduces hallucinations across models.
}
\label{tab:main-results}
\end{table*}

\subsection{Dataset}
\label{sec:datasetting}
We construct a diagnostic dataset from temporal factual QA benchmarks to evaluate temporal evolution and confidence-aware commitment.

\paragraph{Dataset Construction}

We sample from FRESHQA~\footnote{\url{https://github.com/freshllms/freshqa}} and TEMPLAMA~\footnote{\url{https://github.com/google-research/language/tree/master/language/templama}} rather than use them in full, since ChronoRegime requires clear evidence-bounded predictability labels. 
An instance is labeled \(\textsc{Predictable}\) only if a unique future answer is supported by stable persistence, a known boundary, or a deterministic rule/cycle; it is labeled \(\textsc{Unpredictable}\) if the target value depends on external updates or future events outside the bounded evidence.
We refresh the gold annotations to reflect the world state as of \textit{September 2025}. This refresh is used only to correct stale answers in the source benchmarks and to define evaluation targets. We also annotate each instance with an evolution regime, a predictability label, and optional \textit{next\_answer}. 
Details are in Appendix~\ref{app:dataset_examples}.




\paragraph{Dataset Statistics}

The final dataset contains 337 base factual queries: 102 \textit{fast-changing}, 119 \textit{slow-changing}, and 116 \textit{never-changing} facts. Each instance can be instantiated at multiple target times.
Dataset details and additional examples are provided in the Appendix~\ref{app:dataexamples}.

\subsection{Baselines}

\paragraph{Models}
We evaluate CTR on both open-source and closed-source LLMs, including \textbf{LLaMA-3.1-8B-Instruct}~\citep{dubey2024llama}, \textbf{Qwen3-4B-Instruct}~\citep{yang2025qwen3}, \textbf{Mistral-7B-Instruct-v0.3}~\citep{jiang2023mistral7b}, and \textbf{GPT-4o}. 
To avoid temporal leakage, we use model snapshots whose documented knowledge cutoff precedes 2025, while all target queries satisfy \(t^* \geq 2025\).
All models are evaluated in a zero-shot setting with identical prompts and decoding configurations.

\paragraph{Controlled Retrieval Baseline}
We compare CTR with a controlled retrieve-then-reason baseline based on Chain-of-Thought~\citep{Wei2022ChainOT} reasoning. 
The baseline uses the same bounded historical evidence as CTR, allowing us to isolate the effect of CTR's confidence-aware decision mechanism rather than access to historical information alone. 
Details on model checkpoints and the rationale for using bounded evidence instead of open-corpus RAG are provided in Appendix~\ref{app:exp-details}.

\paragraph{Alternative uncertainty and aggregation baselines}
We further compare CTR with three alternative decision strategies on GPT-4o: \textit{Entropy-Gated Selective Prediction}, which abstains based on answer-token entropy; \textit{Heterogeneous Expert Plurality}, which selects the plurality answer among predictions from Qwen3, GPT-4o, LLaMA3.1, and Mistral; and \textit{5-sample Self-Consistency}, which selects the majority answer from five independently sampled GPT-4o generations. These baselines test whether uncertainty estimation or answer agreement alone can account for the reliability gains of CTR.

\subsection{Experimental Setup}

All experiments are conducted with a fixed future time $t^\ast \ge 2025$. 
Unless otherwise specified, CTR probes a historical window of size $W=50$, querying years $\{t^\ast-50,\ldots,t^\ast-1\}$. 
All models are evaluated in a zero-shot setting with a decoding temperature of 0.2 and \textit{max\_new\_tokens} set to 1024. 
Results are averaged over five runs.

Hallucination is measured using a two-stage verification protocol, with details provided in Appendix~\ref{app:hallucination}.

\section{Result Analysis}
\subsection{Overall Performance}

Table~\ref{tab:main-results} compares the Base models, the retrieval-guided baseline, and the proposed CTR framework. Overall, CTR consistently improves answer accuracy and evolution type prediction while substantially reducing hallucination, demonstrating more reliable future reasoning.

CTR outperforms both baselines in answer accuracy across all backbone models, with the largest gains on Slow- and Never-changing facts. On GPT-4o, accuracy on Slow-changing facts increases from 46.2\% to 52.9\% and on Never-changing facts from 79.3\% to 85.3\% compared with controlled retrieval. Importantly, these gains do not degrade performance on Fast-changing facts. GPT-4o with CTR achieves 18.6\% accuracy on Fast facts, exceeding both the Base (16.7\%) and Controlled Retrieval (13.7\%) settings, indicating that CTR selectively commits to predictions rather than suppressing them under uncertainty.

\subsection{Comparison with Alternative Baselines}

As shown in Table~\ref{tab:alternative_baselines}, generic uncertainty or agreement signals alone do not reproduce the overall performance of CTR. Entropy gating achieves the lowest hallucination rate, at 10.4\%, but reduces coverage to 56.4\% and overall accuracy to 41.2\%. Self-consistency substantially improves answer accuracy to 52.8\%, yet its hallucination rate remains high at 26.7\%. Heterogeneous expert aggregation is similarly insufficient, underperforming the single GPT-4o baseline in accuracy while producing more hallucinations.
Controlled Retrieval improves upon direct prediction, but remains behind CTR despite access to the same bounded historical evidence.

Overall, CTR achieves the highest answer accuracy at 53.7\% while maintaining a low hallucination rate of 11.3\%, suggesting that explicit temporal dynamics and commitment regulation provide information beyond generic confidence and answer agreement.
This complements the cross-backbone results in Table~\ref{tab:main-results}, where Controlled Retrieval yields inconsistent gains and can even degrade performance on fast-changing facts.


\begin{table}[t]
\centering
\setlength{\tabcolsep}{6pt}
\begin{tabular}{lcc}
\toprule
Method & Acc. $\uparrow$ & Halluc. $\downarrow$ \\
\midrule
GPT-4o                    & 44.8 & 17.8 \\
Entropy-Gated             & 41.2 & \textbf{10.4} \\
Heterogeneous Plurality   & 42.7 & 28.2 \\
5-sample Self-Consistency & 52.8 & 26.7 \\
Controlled Retrieval      & 47.8 & 13.7 \\
CTR                       & \textbf{53.7} & 11.3 \\
\bottomrule
\end{tabular}
\caption{Comparison with alternative uncertainty- and
agreement-based baselines on GPT-4o. All methods use the same
hallucination evaluation protocol.}
\label{tab:alternative_baselines}
\end{table}


\subsection{Component Ablation}

To examine whether CTR's gains arise from its complete predictability-aware decision process rather than individual heuristic components, we ablate key stages of the pipeline on GPT-4o and Qwen3-4B. Table~\ref{tab:ablation} reports changes relative to the full CTR system in hallucination, coverage, and answer accuracy.
\label{sec:ablation}

\begin{table*}[t]
\centering
\setlength{\tabcolsep}{9pt}
\begin{tabular}{lrrr|rrr}
\toprule
& \multicolumn{3}{c|}{GPT-4o}
& \multicolumn{3}{c}{Qwen3} \\
Ablation
& $\Delta$Hall.$\downarrow$
& $\Delta$Cov.
& $\Delta$Acc.$\uparrow$
& $\Delta$Hall.$\downarrow$
& $\Delta$Cov.
& $\Delta$Acc.$\uparrow$ \\
\midrule
w/o Entropy
& +7.4 & -2.7 & -0.6
& +0.9 & +22.8 & -3.9 \\

w/o Singleton Filtering
& +7.7 & -1.5 & -3.9
& -1.5 & +21.1 & -3.0 \\

w/o Change-Point Modeling
& +5.6 & +0.3 & -0.4
& -1.5 & +24.9 & -4.5 \\

Timeline-only
& +10.1 & +3.0 & -5.5
& +9.2 & +14.2 & -1.4 \\

w/o Commitment Regulation
& +5.9 & +25.2 & -11.6
& +27.6 & +40.9 & -8.3 \\
\bottomrule
\end{tabular}
\caption{Component ablations relative to full CTR.
Positive $\Delta$Hall. indicates increased hallucination.
All values are percentage-point changes.}
\label{tab:ablation}
\end{table*}

The ablations show that the components play complementary roles. On GPT-4o, removing any component increases hallucination, with the largest increase observed for the Timeline-only variant. Removing commitment regulation also substantially expands coverage at the cost of answer accuracy, indicating over-commitment rather than improved reasoning. This effect is more pronounced on Qwen3-4B, where removing commitment regulation increases hallucination by 27.6 points while reducing accuracy by 8.3 points. Although removing singleton filtering or change-point modeling slightly lowers hallucination on Qwen3-4B, both variants substantially increase coverage while decreasing accuracy, indicating a less favorable overall trade-off.

\subsection{Reliability: Hallucination and Abstention}

\begin{figure}[htb]
\centering
\includegraphics[width=1\columnwidth]{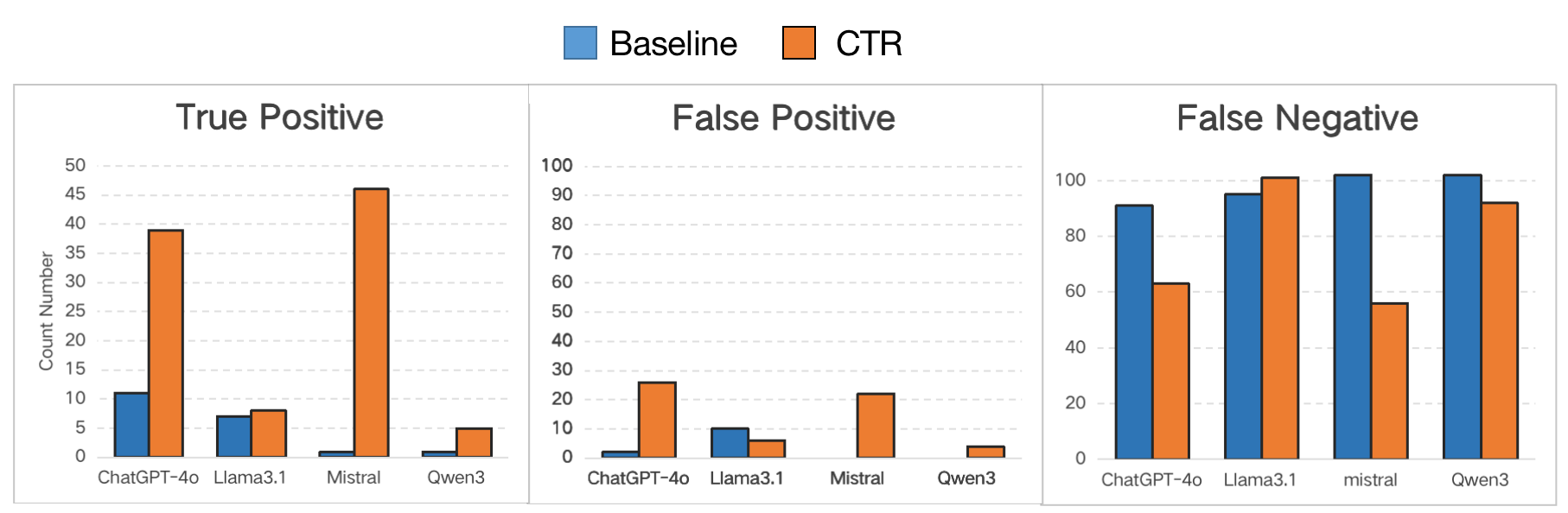}
\caption{
Abstention performance of CTR policy.
}
\label{fig:abstention}
\end{figure}

CTR also yields the largest improvements in reliability. Hallucination rates are consistently lower than both baselines across all models. For Qwen3, hallucination decreases from 56.7\% in the Base setting to 25.5\% under CTR, far below the controlled retrieval result of 54.6\%. GPT-4o exhibits a similarly low hallucination rate of 11.3\%. 

This reliability gain is further reflected in abstention behavior. We treat explicit non-commitment outputs as abstentions. Briefly, TP denotes abstaining on unpredictable instances, FP denotes abstaining on predictable instances, and FN denotes making a concrete prediction on unpredictable instances; formal counting rules are provided in Appendix~\ref{app:abstention}.

As shown in Figure~\ref{fig:abstention}, CTR generally reduces False Negatives on unpredictable cases, for example, from 91 to 63 on GPT-4o and from 102 to 56 on Mistral, while also increasing True Positives. Although False Positives increase for some backbones, this suggests a shift toward a more conservative commitment policy rather than unconditional extrapolation.

Above results support the central ChronoRegime hypothesis: reliable future reasoning depends not only on predicting correctly when evidence is sufficient, but also on refraining from unsupported commitments when predictability is weak.

\subsection{Case Study}
\label{{app:case-study}}
We further analyze representative cases to examine how CTR repairs unreliable future reasoning. 
The case studies show that controlled retrieval reasoning often commits to plausible but weakly supported answers, including unsupported extrapolations, fabricated facts, and noisy change-point interpretations. 
CTR mitigates these failures by aggregating temporal evidence, suppressing isolated low-confidence changes, and committing only when long-term stability supports the prediction. These cases show why retrieval alone is insufficient.
Full case analyses are provided in Appendix~\ref{app:ctr-repair}.

\section{Conclusion}




In this work, we study future-oriented temporal reasoning through the lens of prediction commitment. 
We introduce \textbf{ChronoRegime}, a task that requires models to determine whether historical temporal evidence is sufficient to justify a future prediction. 
Our empirical analysis shows that current LLMs often struggle with this judgment, leading to overconfident commitments under temporal uncertainty.

To address this issue, we propose \textbf{Confidence-Aware Temporal Reasoning (CTR)}, which separates predictability estimation from answer commitment. 
Across four LLMs, CTR consistently improves answer accuracy, reduces hallucination, and provides more transparent grounding for future predictions. 
These findings highlight the importance of modeling predictability, rather than merely extrapolating future answers, for reliable temporal reasoning. More broadly, our results suggest that reliable future reasoning depends on explicitly assessing whether the available temporal evidence is sufficient to support a prediction.

\section*{Limitations}
While CTR demonstrates consistent improvements in future-oriented factual reasoning, several limitations remain.

\textbf{Computational efficiency.}
Our primary experimental setup adopts a rigorous year-by-year probing strategy ($W=50$) to establish a diagnostic upper bound for temporal reasoning performance. We acknowledge that this incurs a linear computational cost ($\mathcal{O}(W)$) relative to the window size. However, the historical recall process is fully parallelizable. Furthermore, for real-world deployment, CTR is designed to function primarily as an offline knowledge curation pipeline, constructing reliable temporal indices once to support low-latency lookups for subsequent queries. Future iterations can further optimize efficiency by implementing adaptive sampling (e.g., binary search) to reduce the probing complexity from linear $\mathcal{O}(W)$ to logarithmic $\mathcal{O}(\log W)$.

\textbf{Uncertainty estimation.}
CTR uses token-level entropy as a proxy for model confidence. While this signal is model-agnostic and easy to compute, it may not fully capture epistemic uncertainty, particularly for models with imperfect probability calibration. Exploring alternative or complementary uncertainty measures remains an important direction for future work.

\textbf{Scope of evaluation.}
Our evaluation focuses on cloze-style temporal factual question answering with discrete time points. While this setting enables controlled analysis of future commitment and abstention, it does not cover more complex scenarios such as open-ended forecasting, interacting events, or long-horizon planning, which remain open challenges for future exploration.

\section*{Ethics Statement}
All work in this paper adheres to the ACL Code of Ethics. 
For the human evaluation of hallucinations, we recruited volunteer annotators and compensated them in accordance with local wage standards. The collected data do not involve any personally identifiable or sensitive information.

The study is conducted in a controlled research setting using released large language models under their respective licenses and usage policies. The proposed framework is intended for methodological analysis of future-oriented factual reasoning. While model-generated outputs may contain inaccuracies, such risks are inherent to large language models and are not the focus of this work.

\section*{Acknowledgments}
This work was supported by the National Key Research and Development Program of China (Grant No. 2024YFC3308101) and the Fundamental Research Funds for the Central Universities (2025CX11021). The authors also thank all reviewers and meta-reviewers for their valuable feedback.



\bibliography{custom}

\clearpage
\appendix

\section{CTR Inference Algorithm}

This appendix provides the detailed inference procedure of Confidence-Aware Temporal Reasoning (CTR). 
CTR is designed to separate two decisions that are often conflated in future-oriented temporal reasoning: 
(i) estimating the temporal predictability regime of the queried fact, and 
(ii) deciding whether the model should commit to a concrete future answer.

Algorithm~\ref{alg:ctr} summarizes the full pipeline. 
Given a query $q$ and a target future time $t^*$, CTR first probes historical states before $t^*$, reconstructs a temporal timeline, detects reliable change signals, induces a predictability regime, and finally applies a commitment policy. 
Unlike unconditional prediction, the final step may return an abstention expression when the inferred evidence does not justify commitment.








\begin{algorithm}[htb]
\caption{Confidence-Aware Temporal Reasoning (CTR)}
\label{alg:ctr}
\small
\begin{algorithmic}[1]
\Require Query $q$, future time $t^\ast$, window size $W$
\Ensure Final output
$\langle r,z,u,\hat{a}_{t^\ast}\rangle$

\Statex \textbf{Historical Recall}
\State $\mathcal{R} \gets \emptyset$
\For{$y=t^\ast-W$ to $t^\ast-1$}
    \State $(a_y,H_y) \gets \mathrm{ProbeModel}(q,y)$
    \State $c_y \gets f(H_y)$
    \State $\mathcal{R}
    \gets \mathcal{R}\cup\{(y,a_y,c_y)\}$
\EndFor

\Statex \textbf{Timeline Reconstruction}
\State $\mathcal{T} \gets \emptyset$
\ForAll{$(y,a_y,c_y)\in\mathcal{R}$}
    \State $v_y \gets a_y$
    \State $q_y \gets
    \operatorname{Clip}_{[0,1]}
    \bigl(c_y(1-\rho_{\mathrm{len}}(y))
    (1-\rho_{\mathrm{rep}}(y))\bigr)$
    \State $\mathcal{T}
    \gets\mathcal{T}\cup\{(y,v_y,q_y)\}$
\EndFor

\Statex \textbf{Change Point Detection}
\State $\widetilde{\mathcal{C}}
\gets \mathrm{ProposeChanges}(\mathcal{T})$
\If{$\widetilde{\mathcal{C}}$ is unavailable or invalid}
    \State $\widetilde{\mathcal{C}}
    \gets \mathrm{DeterministicChanges}(\mathcal{T})$
\EndIf
\State $\mathcal{C}
\gets \mathrm{FilterAndReweight}
(\widetilde{\mathcal{C}},\mathcal{T},
\tau_{\mathrm{single}})$

\Statex \textbf{Predictability Regime Induction}
\State $z \gets
\mathrm{InferEvolutionRegime}(\mathcal{T},\mathcal{C})$
\State $u \gets
\mathrm{InferPredictability}(z,t^\ast,\mathcal{T})$
\State $\mathrm{key\_evidence}
\gets \mathrm{ExtractEvidence}
(\mathcal{T},\mathcal{C},z,u)$
\State $\pi \gets
\langle z,u,\mathrm{key\_evidence}\rangle$

\Statex \textbf{Commitment-Regulated Inference}
\If{$u=\textsc{Predictable}$}
    \State $\hat{a}_{t^\ast}
    \gets
    \underset{a}{\arg\max}\,
    p(a\mid q,t^\ast,\pi)$
\Else
    \State $\hat{a}_{t^\ast}
    \gets \texttt{Abstention}$
\EndIf
\State $r \gets
\mathrm{ExplainDecision}
(q,t^\ast,\pi,\hat{a}_{t^\ast})$

\State \Return
$\langle r,z,u,\hat{a}_{t^\ast}\rangle$
\end{algorithmic}
\end{algorithm}
The commitment policy is intentionally conservative. 
For stable or slowly evolving facts, CTR commits only when the reconstructed timeline provides sufficient support for the future state. 
For highly volatile facts, or when the evidence contains inconsistent or low-confidence transitions, CTR abstains rather than extrapolating unsupported answers. 
Thus, the algorithm operationalizes ChronoRegime as an evidence-bounded decision process rather than an unconditional answer generator.

\section{Detailed Evaluation Metrics and Protocol}
\label{app:eval}

\subsection{Boundary-Probing Metrics}
\label{app:boundary-metrics}

For each boundary-probing instance, we evaluate the model at half-year offsets
around the annotated answer-changing point. We denote the set of pre-change
offsets by \(K^-=\{-6,\ldots,-1\}\) and the set of post-change offsets by
\(K^+=\{0,\ldots,6\}\). Following our implementation, the boundary point
\(k=0\) is included in the post-change region. Let \(a_0\) denote the anchor
answer before the change, \(a_1\) denote the answer after the change, and
\(\hat{a}_k\) denote the model prediction at offset \(k\). For brevity, the
definitions below are written for one instance and then averaged over all
instances.

\paragraph{Pre- and Post-change Accuracy}
We report accuracy separately before and after the change point:
\begin{equation}
\label{eq:pre-post-acc}
\begin{aligned}
\mathrm{PreAcc}
&=
\frac{1}{\lvert K^- \rvert}
\sum_{k \in K^-}
\mathbf{1}\!\left[\hat{a}_k = a_0\right], \\
\mathrm{PostAcc}
&=
\frac{1}{\lvert K^+ \rvert}
\sum_{k \in K^+}
\mathbf{1}\!\left[\hat{a}_k = a_1\right].
\end{aligned}
\end{equation}
PreAcc measures whether the model recovers the anchor answer while it is still
valid, whereas PostAcc measures whether the model adapts after the anchor
answer expires.

\paragraph{Change-Aware Success}
CAS measures whether the model succeeds on both sides of the boundary for the
same instance. We count an instance as successful only when the model answers
correctly immediately before and after the change point:
\begin{equation}
\label{eq:cas}
\mathrm{CAS}
=
\mathbf{1}\!\left[
\hat{a}_{-1}=a_0
\land
\hat{a}_{1}=a_1
\right].
\end{equation}
The reported CAS score is the average of this indicator over all instances.

\paragraph{Post-change Persistence}
Post-change Persistence Rate (Post-PR) measures how often the model continues
to output the expired anchor answer in the post-change region:
\begin{equation}
\label{eq:post-pr}
\mathrm{Post\mbox{-}PR}
=
\frac{1}{\lvert K^+ \rvert}
\sum_{k \in K^+}
\mathbf{1}\!\left[\hat{a}_k = a_0\right].
\end{equation}
A lower Post-PR indicates that the model is less likely to persist with an
outdated answer after the boundary.

\paragraph{Wrong Concrete Commitment}
WCCR measures incorrect non-abstaining predictions in the post-change region.
Let \(\mathcal{U}(\hat{a}_k)\) indicate that the prediction is an abstention or
uncertainty response, such as "unknown", "uncertain", or "cannot be
determined". Then
\begin{equation}
\label{eq:wccr}
\mathrm{WCCR}
=
\frac{1}{\lvert K^+ \rvert}
\sum_{k \in K^+}
\mathbf{1}\!\left[
\neg \mathcal{U}(\hat{a}_k)
\land
\hat{a}_k \ne a_1
\right].
\end{equation}
Thus, WCCR captures cases where the model makes a concrete commitment after the
change point but commits to the wrong answer.

\paragraph{Entropy Shift}
Finally, we report the change in answer entropy across the boundary:
\begin{equation}
\label{eq:entropy-shift}
\begin{aligned}
\Delta H
&= H_{\mathrm{post}} - H_{\mathrm{pre}}, \\
H_{\mathrm{pre}}
&=
\frac{1}{\lvert K^- \rvert}
\sum_{k \in K^-} H_k, \\
H_{\mathrm{post}}
&=
\frac{1}{\lvert K^+ \rvert}
\sum_{k \in K^+} H_k .
\end{aligned}
\end{equation}
Here, \(H_k\) is the answer entropy at offset \(k\). A positive \(\Delta H\)
indicates that the model becomes more uncertain after crossing the boundary.
\subsection{Answer Correctness Criteria}
\label{app:correctness}

\paragraph{Strict Correctness}
We primarily use strict correctness to evaluate whether the model
identifies the exact factual state. This serves as our main accuracy
criterion.

\paragraph{Lenient Post-change Correctness}
For future-oriented queries, strict evaluation may penalize
reasonable caution. We therefore introduce a lenient criterion
for post-change probes. Let $\mathcal{A}_{\mathrm{abs}}$ denote
the set of explicit abstention or uncertainty expressions,
including outputs containing "Unknown", "Uncertain", or "TBD." For $k \in K^+$, a response is considered correct if
it either matches the post-change answer $a_1$ or belongs to
$\mathcal{A}_{\mathrm{abs}}$; for $k \in K^-$, we retain the
strict correctness criterion:


\begin{equation}
\label{eq:lenient-correctness}
\begin{aligned}
&\mathrm{Correct}^{\mathrm{lenient}}_k \\
&\quad =
\begin{cases}
\mathrm{Correct}^{\mathrm{strict}}_k,
& k \in K^-, \\[2pt]
\mathbf{1}\!\left[
\hat{a}_k \in \{a_1\} \cup
\mathcal{A}_{\mathrm{abs}}
\right],
& k \in K^+.
\end{cases}
\end{aligned}
\end{equation}

This metric distinguishes between correctable epistemic uncertainty and
confident hallucinations.

\subsection{Uncertainty Estimation}
\label{app:uncertainty}

We quantify model confidence using token-level entropy derived from the
decoding process. Let $\mathcal{T}_k$ denote the set of token indices
corresponding to the generated answer $\hat{a}_k$. The average answer
entropy for an instance at time offset $k$ is defined as:

\begin{equation}
\label{eq:answer-entropy}
H_k
=
\frac{1}{|\mathcal{T}_k|}
\sum_{j \in \mathcal{T}_k}
\mathcal{H}
\left(
p_{\theta}(\cdot \mid x_{<j}, x_k)
\right),
\end{equation}

where $p_{\theta}$ is the model's predictive distribution and
$\mathcal{H}(\cdot)$ denotes Shannon entropy. The entropy shift reported
in Table~\ref{tab:preliminary} is computed according to
Equation~\ref{eq:entropy-shift}.

\subsection{Abstention Evaluation}
\label{app:abstention}

To evaluate whether a model knows when not to commit, we additionally
report abstention-oriented statistics on unpredictable instances. Let
$u_i = 1$ indicate that instance $i$ is unpredictable and $u_i = 0$
indicate that it is predictable. We define TP, FP, and FN as:

\begin{equation}
\label{eq:abstention-stats}
\begin{aligned}
\mathrm{TP}
&=
\sum_i
\mathbf{1}\!\left[
u_i = 1
\land
\hat{a}_i \in \mathcal{A}_{\mathrm{abs}}
\right], \\
\mathrm{FP}
&=
\sum_i
\mathbf{1}\!\left[
u_i = 0
\land
\hat{a}_i \in \mathcal{A}_{\mathrm{abs}}
\right], \\
\mathrm{FN}
&=
\sum_i
\mathbf{1}\!\left[
u_i = 1
\land
\hat{a}_i \notin \mathcal{A}_{\mathrm{abs}}
\right],
\end{aligned}
\end{equation}
where $i$ indexes evaluation instances.

Thus, TP denotes abstaining on an unpredictable instance, FP denotes
abstaining on a predictable instance, and FN denotes making a concrete
prediction on an unpredictable instance.

\section{Prediction Flip Curves around Factual Change Points}
\begin{figure*}[htb]
\centering
\includegraphics[width=\textwidth]{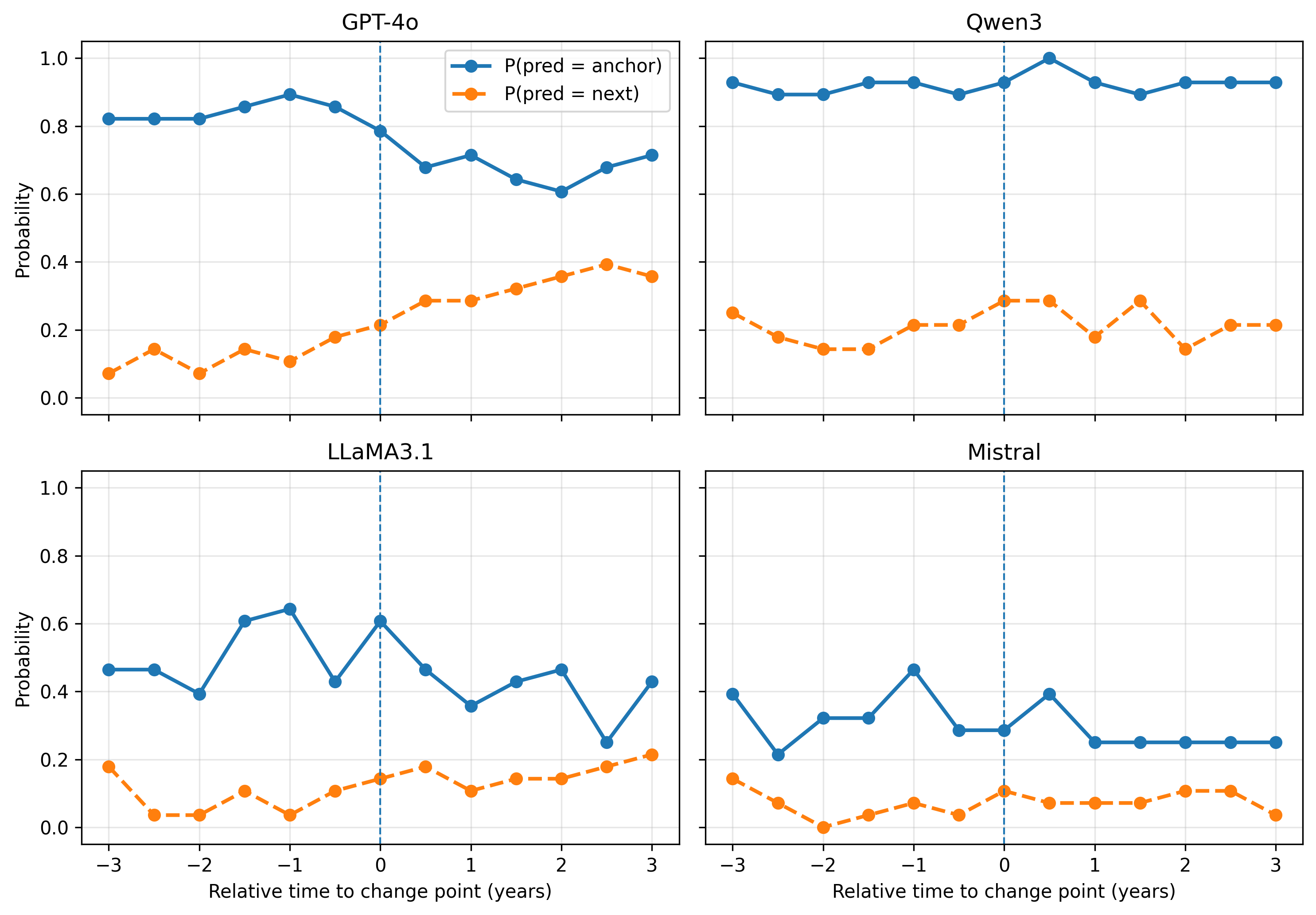}
\caption{
Prediction flip curves around factual change points. 
The blue curve shows the probability of predicting the pre-change anchor answer, while the orange curve shows the probability of predicting the post-change answer. 
A boundary-aware model should reduce anchor-answer predictions and increase post-change predictions after the change point.
}
\label{fig:clip}
\end{figure*}

Figure~\ref{fig:clip} plots model predictions around factual change points. 
For each relative time offset, we report the probability that the model predicts the pre-change anchor answer and the probability that it predicts the post-change answer. 
This figure complements the preliminary study by visualizing whether models actually switch from the old answer to the new answer after the temporal boundary.

Across models, the probability of predicting the anchor answer often remains high even after the change point, while the probability of predicting the post-change answer increases only weakly. 
This provides additional evidence of outdated persistence and weak boundary-sensitive adaptation.

\begin{table*}[t]
\centering
\small
\renewcommand{\arraystretch}{0.8} 
\setlength{\tabcolsep}{6pt}
\begin{tabular*}{\textwidth}{@{\extracolsep{\fill}}
p{2.6cm}      
c             
c             
p{5.0cm}      
p{5.0cm}      
}
\toprule
\textbf{Case} &
\textbf{Model} &
\textbf{Gold} &
\textbf{Controlled Retrieval} &
\textbf{Confidence-Aware TR (CTR)} \\
\midrule

\raggedright
The number of books published by Shoshana Zuboff &
GPT-4o &
3 &
\textbf{Answer:} \textcolor{red}{4} \newline
\textbf{Reasoning:}
Assumes a slow publication rate without citing concrete historical evidence. &
\textbf{Answer:} \textcolor{teal}{3} \newline
\textbf{Reasoning:}
"The number of books remained stable at 3 from 2018 to 2024. Given the \textit{Slow-changing} pattern, it is likely unchanged in 2025."
\\ \midrule

\raggedright
Children of Leonardo DiCaprio &
Qwen3 &
0 &
\textbf{Answer:} \textcolor{red}{1} \newline
\textbf{Reasoning:}
"Leonardo DiCaprio has been publicly known to have one child, a daughter named Matilda, born in 2004."(\textit{Hallucination}) &
\textbf{Answer:} \textcolor{teal}{0} \newline
\textbf{Reasoning:}
"The evidence indicates a stable period from 1978 to 2024 with no children."
\\

\bottomrule
\end{tabular*}
\caption{Two representative case studies comparing the controlled retrieve-then-reason baseline with our CTR framework. The baseline exhibits speculative extrapolation (Case 1) and factual hallucination (Case 2), whereas CTR produces more reliable predictions through structured temporal grounding.}
\label{tab:case-study}
\end{table*}

\section{Experimental Details}
\label{app:exp-details}


\paragraph{Model checkpoints.}
We use LLaMA-3.1-8B-Instruct~\footnote{https://huggingface.co/meta-llama/Llama-3.1-8B-Instruct}, Qwen3-4B-Instruct~\footnote{https://huggingface.co/Qwen/Qwen3-4B-Instruct-2507}, and Mistral-7B-Instruct-v0.3 from Hugging Face, and GPT-4o through the official API. 
All models are evaluated without browsing, tools, open-corpus retrieval, or access to refreshed gold annotations.

\paragraph{Controlled retrieval baseline.}
We do not use open-corpus RAG as the baseline, since it would introduce retrieval-specific confounds such as corpus construction, chunking, retriever quality, and temporal leakage from post-hoc updated documents. 
Both CTR and the baseline operate on the same bounded historical evidence, allowing us to focus on reasoning and commitment behavior rather than retrieval pipeline design.

\paragraph{Implementation.}
Experiments are run on a Linux server equipped with 4 NVIDIA A100 GPUs with 80GB memory each.

\section{Case Studies and Confidence-Aware Repair Analysis}
\label{app:ctr-repair}

We provide additional qualitative analyses to illustrate how CTR improves reliability in future-oriented temporal reasoning. 
The cases cover three representative failure modes of controlled retrieval reasoning: unsupported extrapolation, factual hallucination, and spurious temporal change detection.

\subsection{Unsupported Extrapolation and Factual Hallucination}

Table~\ref{tab:case-study} compares the controlled retrieval baseline with CTR on two representative cases. 
These cases highlight how controlled retrieval reasoning can produce plausible but weakly supported future answers, while CTR regulates commitment based on temporal evolution evidence.

\paragraph{Unsupported extrapolation.}
In the case concerning Shoshana Zuboff's publications, the retrieval-guided baseline predicts an increase in the number of books without citing concrete historical evidence, reflecting a tendency to assume change by default. 
CTR, in contrast, evaluates the temporal reliability of historical recall and identifies a long stable period with consistently high confidence. 
Based on this inferred slow-changing evolution pattern, CTR conservatively maintains the historical value for 2025.

\paragraph{Factual hallucination.}
In the Leonardo DiCaprio case, the retrieval-guided baseline fabricates a specific but non-existent detail, namely ``a daughter named Matilda'', showing that LLMs can generate coherent yet unsupported narratives without explicit grounding. 
CTR avoids this error by aggregating decades of consistent historical evidence. 
The confidence-aware mechanism filters out low-probability generative noise, allowing the model to conclude that the factual state remains ``Zero''.

\subsection{Confidence-Aware Repair under Noisy Historical Recall}

\begin{figure*}[htb]
\centering
\includegraphics[width=\textwidth]{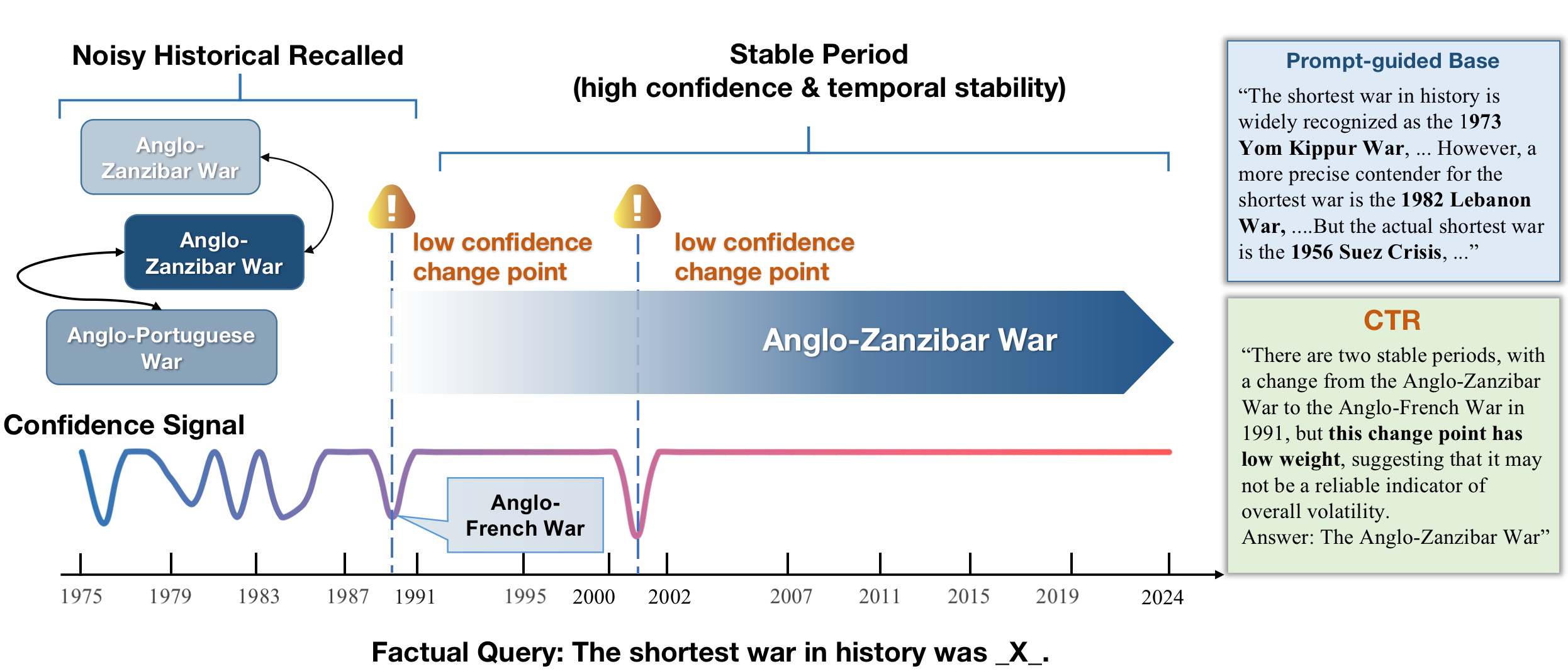}
\caption{
An illustrative case showing how CTR suppresses low-confidence change points and commits to future predictions only when long-term temporal stability is observed.
} 
\label{fig:case}
\end{figure*}

Figure~\ref{fig:case} illustrates a representative case for the query "What was the shortest war in history?", highlighting how CTR preserves prediction reliability under noisy historical recall.

In early years, historical recall exhibits substantial fragmentation, alternating between multiple candidates such as the Anglo-Zanzibar War and the Anglo-Portuguese War. 
Although some recalled answers appear with high instantaneous confidence, the overall confidence signal fluctuates sharply, making these early predictions unreliable.

Around 1991, a deviation to the Anglo-French War is detected as a potential change point. 
However, this change point receives low weight because it is isolated and lacks sustained support in subsequent years. 
CTR therefore treats it as a spurious fluctuation rather than genuine factual evolution.

From 1992 onward, the Anglo-Zanzibar War forms a long stable period with consistently high confidence and low variance, establishing a reliable temporal anchor. 
Based on this stability, CTR commits to the Anglo-Zanzibar War as the future answer.

In contrast, the retrieval-guided baseline produces a fluent but hallucinated reasoning chain by enumerating multiple unrelated conflicts without temporal grounding. 
This comparison demonstrates that CTR enables conservative future commitment by suppressing unreliable change signals and prioritizing long-term temporal stability.

Overall, these cases show that CTR acts as a reliability layer for future-oriented temporal reasoning: it reduces unsupported extrapolation, avoids factual hallucination, and prevents isolated noisy recalls from being misinterpreted as meaningful temporal changes.

\section{Reasoning Hallucination Judgement}
\label{app:hallucination}

To ensure the reliability of our hallucination evaluation, we employ a rigorous two-stage verification pipeline, combining broad automated screening with fine-grained human adjudication.

\paragraph{Stage 1: Strict Automated Screening.}
First, we utilize GPT-5.1 as an independent judge to perform an initial screening of all reasoning traces. The judge is instructed to flag instances based on three strict criteria:
\begin{enumerate}
    \item \textbf{Fabrication:} Citing specific events, names, or data points that do not exist (e.g., inventing a "2025 release date" for a cancelled movie).
    \item \textbf{Unsupported Extrapolation:} Making definitive claims about volatile future states without citing stable historical patterns or official announcements.
    \item \textbf{Logical Contradiction:} Deriving a conclusion that contradicts the cited historical evidence within the trace.
\end{enumerate}
This stage acts as a high-recall filter to identify potential hallucinations ($\mathrm{Hallucination}_{i,k} = 1$).

\paragraph{Stage 2: Expert Human Adjudication.}
To eliminate false positives from the automated judge and ensure gold-standard quality, we conduct a secondary human review. 
This process is performed by expert annotators familiar with the temporal reasoning task. The human review protocol involves:
\begin{itemize}
    \item \textbf{Fact Verification via Search:} For every flagged instance, annotators are required to verify the validity of the generated claims against external search engines (e.g., Google Search) to confirm whether the "hallucinated" detail is indeed false or merely obscure.
    \item \textbf{Logic and Grounding Check:} Annotators assess whether the reasoning logically follows from the provided context. A trace is marked as valid only if the prediction is grounded in verifiable patterns (e.g., "consistent 4-year cycle") rather than spurious correlations.
    \item \textbf{Final Adjudication:} In cases of disagreement between the automated judge and human assessment (e.g., the model correctly inferred a complex fact that GPT-5.1 mistook for hallucination), the human judgment prevails.
\end{itemize}
This human-in-the-loop refinement ensures that our reported hallucination rates reflect genuine generative failures rather than evaluation artifacts.

We emphasize that hallucination rates are used for controlled relative comparison across methods under identical judging conditions, rather than as absolute measures of factuality.
\label{app:entropy}

\section{Dataset Selection and Annotation Details}
\label{app:dataset_examples}

\begin{table*}[h]
\centering
\small
\renewcommand{\arraystretch}{1.4} 
\setlength{\tabcolsep}{4pt}
\begin{tabular*}{\textwidth}{@{\extracolsep{\fill}} p{1.8cm} p{5.0cm} c p{3.2cm} p{2cm}}
\toprule
\textbf{Fact Type} & 
\textbf{Query ($q_i$)} & 
\textbf{Ref. Time ($t_0^i$)} & 
\textbf{Current Answer ($a_0^i$)} & 
\textbf{Next Answer ($a_1^i$)} \\
\midrule

\textit{Never} & 
\_X\_ won the 2020 Formula 1 world driver's championship. & 
2022 & 
Lewis Hamilton & 
\texttt{N/A} \\

\textit{Never} & 
The real name of the Unabomber was \_X\_. & 
2022 & 
Theodore John Kaczynski & 
\texttt{N/A} \\
\midrule

\textit{Slow} & 
The Winter Olympics were held most recently in \_X\_. & 
2022 & 
Beijing, China & 
Italy \\

\textit{Slow} & 
The next leap year is \_X\_. & 
2024 & 
2028 & 
2032 \\
\midrule

\textit{Fast} & 
The most recent movie in the Marvel Cinematic Universe is \_X\_. & 
2025 & 
The Fantastic Four: First Steps & 
Spider-Man: Brand New Day \\

\textit{Fast} & 
King Gizzard's most recent studio album is \_X\_. & 
2025 & 
Phantom Island & 
\texttt{N/A} \\

\bottomrule
\end{tabular*}
\caption{
Representative dataset examples. 
The Next Answer field is used only for evaluation and is not provided to the model. 
\texttt{N/A} indicates that no annotated future transition answer is provided; its interpretation depends on the evolution type.
}
\label{tab:dataset_examples}
\end{table*}

\subsection{Predictability Labeling Protocol}
Predictability labels are assigned at the instance level under the bounded-evidence setting. Let \(E_i\) denote the historical temporal evidence available for instance \(i\), and \(t_i^*\) the queried future time. We label \(i\) as \(\textsc{Predictable}\) if \(E_i\) supports a unique answer at \(t_i^*\) through one of the following patterns: 

(i) stable persistence, where the factual state remains unchanged across the relevant horizon; 

(ii) a known temporal boundary, where the validity interval of the current or next state is specified;

(iii) a deterministic rule or cycle, such as calendar-based recurrence. 
If the target value depends on external announcements, future events, or updates not available in \(E_i\), the instance is labeled \(\textsc{Unpredictable}\).

This criterion is independent of the evolution regime annotation. In the final diagnostic set, FAST instances are selected when they meet the \(\textsc{Unpredictable}\) criterion, whereas SLOW and NEVER instances are selected when they meet the \(\textsc{Predictable}\) criterion. Thus, the observed type-label mapping is induced by our selection protocol and should not be interpreted as a universal claim that all FAST facts are unpredictable or all SLOW/NEVER facts are predictable.

\subsection{Boundary Cases}
A FAST-changing fact can be labeled \(\textsc{Predictable}\) if the future value is already fixed by bounded evidence, such as an officially announced schedule. Such cases are not included in the FAST subset of our diagnostic benchmark. Conversely, a SLOW-changing fact can be labeled \(\textsc{Unpredictable}\) if the queried time lies beyond a known validity boundary and no successor state is supported by the evidence. These cases are excluded or relabeled during annotation. This filtering ensures that the benchmark evaluates evidence-bounded predictability rather than raw volatility alone.

\subsection{Gold Refresh}
The gold refresh date is \textit{September 2025}. This means that stale answers inherited from FRESHQA or TEMPLAMA are re-verified and corrected against the world state at that date. The refreshed gold answer is used only as an evaluation label. It is not part of the bounded historical evidence and is never shown to the model. For rule-governed instances, such as calendar questions, the label is computed from the target time \(t^*\).
\subsection{Next Answer}
The \textit{next\_answer} field is not the predictability label. It is recorded only when a future state is externally verifiable and useful for evaluation. For NEVER facts, \textit{N/A} means that no future transition is annotated because the current answer is expected to remain stable. For FAST facts, \textit{N/A} means that no stable or justified future target is assumed under bounded evidence.


\section{Dataset Examples}
\label{app:dataexamples}
Table~\ref{tab:dataset_examples} presents representative examples from our evaluation dataset. 
Each instance is annotated with its evolution type, cloze-style query, reference time, current answer, and an optional next-answer field. The structure strictly follows the formal definition provided in Section~3, where each instance is annotated as a tuple $e_i = \langle q_i, t_0^i, a_0^i, t_c^i, a_1^i \rangle$.
The transition time \(t_c^i\) and next answer \(a_1^i\)are used for evaluation and analysis only, and are not provided to the model during probing or inference.

Specifically, the table columns correspond to:
\begin{itemize}
    \item \textbf{Fact Type}: The evolution category.
    \item \textbf{Query ($q_i$)}: The cloze-style factual query.
    \item \textbf{Ref. Time ($t_0^i$)}: The latest reference time at which the fact is known to be valid (corresponding to the \texttt{effective\_year} in the raw data).
    \item \textbf{Current Answer ($a_0^i$)}: The correct answer valid at time $t_0^i$.
    \item \textbf{Next Answer ($a_1^i$)}: The verifiable future answer after the change point. This is set to \texttt{N/A} for \textit{Never-changing} facts or \textit{Fast-changing} facts where the future state is highly volatile or yet to be determined.
\end{itemize}

The \textit{Current Answer} field denotes the factual answer verified at the reference time. 
The \textit{Next Answer} field is used only for evaluation and analysis when a future answer is well-defined and verifiable. 
It is not provided to the model during inference.

Importantly, \texttt{N/A} in the Next Answer field should not be interpreted uniformly as an unpredictable future state. 
For \textsc{Never}-changing facts, \texttt{N/A} means that no future transition is annotated because the current answer is expected to remain stable. 
For \textsc{Fast}-changing facts, \texttt{N/A} indicates that no stable or justified future answer is assumed under the available evidence. 
Thus, evaluation decisions depend on both the evolution type and the availability of a verifiable future answer.

\end{document}